# Causality in Bayesian Belief Networks

Marek J. Druzdzel
Carnegie Mellon University
Department of Engineering
and Public Policy
Pittsburgh, PA 15213
marek+@cmu.edu

Herbert A. Simon
Carnegie Mellon University
Department of Psychology
Pittsburgh, PA 15213
has+@a.gp.cs.cmu.edu

## Abstract

We address the problem of causal interpretation of the graphical structure of Bayesian belief networks (BBNs). We review the concept of causality explicated in the domain of structural equations models and show that it is applicable to BBNs. In this view, which we call *mechanism-based*, causality is defined within models and causal asymmetries arise when mechanisms are placed in the context of a system. We lay the link between structural equations models and BBNs models and formulate the conditions under which the latter can be given causal interpretation.

## 1 INTRODUCTION

Although references to causality permeate everyday scientific practice, the notion of causation has been one of the most controversial subjects in the philosophy of science. Hume's critique that causal connections cannot be observed, and therefore have no empirical basis, strongly influenced the empiricist framework and refocused the concept of causality to scientific models as opposed to reality. A strong attack on causality was launched in the beginning of this century by Bertrand Russel, who, observing the developments in areas of physics such as gravitational astronomy, argued that causality is a "relic of bygone age," for which there is no place in modern science.[1] Philosophical attempts to capture the meaning of causation and reduce it to a theoretically sound and meaningful term succeeded only in part. Although they exposed and clarified several important issues related to the concept of causality and its use in science, no known philosophical definition of causation is free from objections or examples in which it seems to fail. This has created an atmosphere of suspicion towards the very concept. It is, therefore, not surprising that many scientists are rather careful in using the term causality, preferring neutral mathematical terms like "functional relation" or "interdependence." Still, capturing the asymmetries implied by causality seem to be an inherent part of scientific practice.

The confusion about the meaning of causality is clearly seen in the domain of probabilistic and decision-theoretic models based on Bayesian belief networks (BBNs) (Pearl, 1988) and influence diagrams (Howard & Matheson, 1984). On one hand, the directionality of arcs brings into a model asymmetric relations among variables, which some theoreticians have associated with cause-effect relations (e.g., Pearl (1988), Lauritzen and Spiegelhalter (1984)). In causal discovery work (Spirtes *et al.* (1993), Pearl and Verma (1991)), the relation between causality and probability is bound by a set of axioms that allow for causal inference. However, the faithfulness (or minimality) assumption in causal discovery is too restrictive for a definition of causality: it is possible for a causal graph to produce an unfaithful probability distribution. Some theoreticians pointed out that BBNs are simply a mathematical formalism for representing explicitly dependences and independences, and that there is no inherent relation of directed arcs with causality in the formalism (e.g., Howard and Mathesson (1984)). After all, the arcs can be reversed simply by application of Bayes' rule, whereas causality cannot.

There seems to be little doubt that the notion of causality is useful in probabilistic models. There is strong evidence that humans are not indifferent to causal relations and often give causal interpretation to conditional probabilities in the process of eliciting conditional probability distributions (Tversky & Kahneman, 1977). Henrion (1989) gives an appealing practical example when a little reflection on the causal structure of the domain helps a domain expert to refine the model. Discovery of the fact that an early version of a model violates conditional independence of variables (a consequence of the Markov property) leads the expert to realize that there is an additional intermediate node in the causal structure of the system and subsequently to refine the model. The probabilistic consequences of the causal structure, in terms of the pattern of dependences, are so strong that an expert seek-ing to fulfill the Markov condition, in fact, often ends up looking for the right causal model of the domain. Even those holding the strict "probabilistic influence" view admit that experts often construct influence diagrams that correspond to their causal models of the system (Shachter & Heckerman, 1988). The same can be said about the user interfaces to decision support systems: having a model that represents causal interactions aids in explaining the reasoning based on that model. Experiments with rule-based expert systems, such as Mycin, have indicated that diagnostic rules alone are not sufficient for generating understandable explanations and that at some level a model incorporating the causal structure of the domain is needed (Clancey, 1983; Wallis & Shortliffe, 1984).

---

[1] He later retreated from this extreme view, recognizing the fundamental role of causality in physics.

Usefulness for human interfaces is not the only reason for capturing causality in probabilistic models. As long as the only goal of using a model is prediction of a probability distribution given some evidence (this is the case in typical diagnostic tasks), the notion of causality is technically not useful. Consider, for example, a model consisting of two variables: *weather* and *barometer*. Given the outcome of one of the variables, we can do extremely well in predicting the probability distribution of the other. The problems will begin when we want to predict the effect of a "change in structure" of our system, i.e., the change in some mechanism in the system through an external intervention.[2] Without knowing the direction of the causal relation between the *weather* and the *barometer*, we cannot tell whether a manual manipulation of the *barometer* will affect the *weather*. If this problem sounds unrealistic, consider a public policy decision regarding, for example, banning products that are high in cholesterol, given their observed probabilistic association with heart disease. Without the information on the causal interaction between cholesterol intake and cholesterol blood level, and then cholesterol blood level and heart disease, we can at best predict the effect of our policy decision on the amount of cholesterol intake but not its ultimate effect on heart disease. The effect of a structural change in a system cannot be induced from a model that does not contain causal information. Having the causality right is crucial for any policy making.

One never deals with changes in structure in the domain of decision analysis — all policy options and instruments that are expected to affect a system are explicitly included in the decision model. Whatever causal knowledge is necessary for building this model is assumed to be possessed by the decision maker, and is captured in the conditional probability distributions in the model. The decision maker is assumed to know that, for example, manipulating the barometer will not affect the weather. The problem is pushed away from the formalism to the interaction between the decision analyst and the decision maker and, effectively, since reference to causality seems to be unnecessary in decision models, decision theorists and decision analysts can afford to deny any connection between directionality of arcs and causality.[3]

While one can get away with such a finesse in decision analysis, causal knowledge needs to be made explicit in situations where the human–model loop is less tight. The ability to predict the effect of changes in structure is important for intelligent decision support systems that autonomously generate and evaluate various decision options (intelligent planners). To be able to perform this task, they need a way to compute the effect of imposing values or probability distributions on some of the variables in a model. This can be done only if the model contains information about the causal relations among its variables.

What, in our opinion, deters decision theorists from explicitly including causality in their models is a lack of a theoretically sound and meaningful representation of causality within probabilistic models. In this paper, we propose that the meaning of causality provided by Simon (1953) within structural equations models is extendible to BBNs and can fill the existing niche. In short, the answer given in this paper is that BBNs, taken as a pure formalism, indeed have nothing in them that would advocate a causal interpretation of the arcs. Probabilistic independences in themselves do not imply a causal structure and a causal structure does not necessarily imply independences. To give the arcs a causal interpretation, additional assumptions are necessary. Those researchers who give BBNs the interpretation of causal graphs are justified in doing so in as much as these assumptions hold in their graphs. We make these assumptions explicit, and we hope that this will contribute to reconciling the two views of BBNs.

The remainder of this paper is structured as follows. We first review the principles underlying structural equations models and Simon's procedure for extracting the causal ordering from such models (Section 2). Then, in Section 3, we demonstrate that any BBN model can be represented by a simultaneous equations model with hidden variables. Using this result, in combination with the assumption of acyclicity of BBNs, we outline the conditions under which a BBN can be given a causal interpretation.

---

[2] This problem has been known in philosophy as the "counterfactual conditional," as it involves evaluation of a counterfactual predicate: "if $A$ were true, then $B$ would be the case." See Simon and Rescher (1966) for a discussion of the role of causality in counterfactual reasoning.

[3] Decision nodes in influence diagrams are a clear exception: both incoming and outgoing arcs can be given a causal interpretation. The arcs coming into a decision node denote relevant information, known prior to the decision, that has impact on the decision (i.e., causes the decision maker to choose different options). The arcs coming out of a decision node stand for manipulation of the model's variables or, if they go to the value node, the impact on the overall utility of the decision.

# 2  SIMULTANEOUS EQUATIONS MODELS

Pieces of the real world that can reasonably be studied in isolation from the rest of the world, are often called *systems*. Systems can be natural (the solar system) or artificial (a car), can be relatively simple (a pendulum) or extremely complex (the human brain). Although systems are always interlocked with the rest of the world, one can make a strong philosophical argument that they usually consist of strongly interconnected elements, but that their connections with the outside world are relatively weak (Simon, 1969). This property allows them to be successfully studied in isolation from the rest of the world.

Abstractions of systems, used in science or everyday thinking, are often called *models*. There is a large variety in the complexity and rigor of models: there are informal mental models, simple black-box models, and large mathematical models of complex systems involving hundreds or thousands of variables. A common property of models is that they are simplifications of reality. By making simplifying assumptions, a scientist often makes it practically possible to study a system but, on the other hand, automatically changes the focus of his or her work from reality to its model.

One way of representing models is by sets of simultaneous equations, where each equation describes a functional relation among a subset of the model's variables. Such models are usually self-contained in the sense that they have as many equations as they have variables and, by virtue of the fact that they describe an existing system, have at least one solution and at most a denumerably infinite set of solutions. Often, equations contain so called error variables, variables that are exogenous and usually independent by assumption, and which represent the joint effect of other variables that one is unwilling or unable to specify.

A generic form of an equation that will be used throughout this paper is

$$f(x_1, x_2, \ldots, x_n, \mathcal{E}) = 0 , \qquad (1)$$

where $f$ is some algebraic function, its arguments $x_1$, $x_2$, ..., $x_n$ are various system variables, and $\mathcal{E}$ is an error variable. This form is usually called an *implicit function*. In order to obtain a variable $x_i$ ($1 \leq x_i \leq n$) as a function of the remaining variables, we must solve the equation (1) for $x_i$. We say that the function

$$x_i = g(x_1, x_2, \ldots, x_{i-1}, x_{i+1}, \ldots, x_n, \mathcal{E}) \qquad (2)$$

found in this way is defined *implicitly* by (1) and that the solution of this equation gives us the function explicitly. Often, the solution can be stated explicitly in terms of elementary functions. In other cases, the solution can be obtained in terms of an infinite series or other limiting process; that is, one can approximate (2) as closely as desired.[4]

---
[4]Some implicit functions have no solutions in specified

## 2.1  STRUCTURAL EQUATIONS

As, in most mathematical formalisms, certain classes of transformations are solution-preserving, any model of a system can have many forms, equivalent with respect to the set of solutions. Each such form is an algebraic transformation of some other form.

For each natural system, there is one form that is specially attractive because of its relation to the causal structure of the system. It is a form in which each equation is *structural*, in the sense of describing a conceptually distinct, single mechanism active in the system. An example of a structural equation might be $f = ma$, where $f$ stands for a force active in the system, $m$ for the mass of a system component, and $a$ the acceleration of that component. Another equation might be $p = C_1 - C_2 d$, where $p$ stands for the price of a good, $d$ stands for the demand for that good, and $C_1$ and $C_2$ are constants.

The concept of a structural equation is not mathematical, but semantic. Consequently, there is no formal way of determining whether an equation is structural or not. Structural equations are defined in terms of the mechanism that they describe. The notion of a mechanism can be operationalized by providing a procedure for determining whether the mechanism is present and active or not. Sometimes a mechanism is visible and tangible. One can, for example, expose the clutch of a car and even touch the plates by which the car's engine is coupled with the wheels. One can even provide a graphic demonstration of the role of this mechanism by starting the engine and depressing the clutch pedal. Often, especially in systems studied in social sciences, a mechanism is not as transparent. Instead, one often has other clues or well-developed and empirically tested theories of interactions in the system that are based on elementary laws like "no action at a distance" or "no action without communication" (Simon, 1977, page 52). Structural equations may be formed entirely on the basis of a theory or consist of principles derived from observations, knowledge of legal and institutional rules restricting the system (such as tax schedules, prices, or pollution controls), technological knowledge, physical, chemical, or social laws. They may, alternatively, be formed on a dual basis: a theory supported by systematically collected data for the relevant variables.

A variable is considered *exogenous* to a system if its value is determined outside the system, either because we can control its value externally (e.g., the amount of taxes in a macro-economic model) or because we believe that this variable is controlled externally (like the weather in a system describing crop yields, market prices, etc.). Equations specifying the values of exogenous variables form a special subclass in an struc-

---
domains — the equation $f(x, y) = x^2 + y^2 + 1 = 0$, for example, is satisfied by no real values. All implicit functions referred to in this paper are assumed to have solutions.

tural equations model. An equation belonging to this subclass usually sets the value of a system's variable to a constant, expressing the fact that the value of that variable is determined outside the modeled system, hence, the variable is exogenous to the system.

Often, the core of a simultaneous structural equations model of a natural system will contain fewer equations than variables, hence, forming a system that is underdetermined. Only the choice of exogenous variables and the subsequent addition of equations describing them makes the system self-contained and solvable for the remaining (endogenous) variables. Whether a variable is exogenous or endogenous depends on the point of view on the system that one is describing. The boundaries that one decides to put around the system and one's ability to manipulate the system's elements are crucial for which variables are exogenous and which are endogenous in that system. A variable that is exogenous in a simple system may become endogenous in a larger system.

In a structural equation describing a mechanism $\mathcal{M}$

$$f_\mathcal{M}(x_1, x_2, x_3, \ldots, x_n, \mathcal{E}) = 0,$$

the presence of a variable $x_i$ means that the system's element that is denoted by $x_i$ directly participates in the mechanism $\mathcal{M}$. If a variable $x_j$, in turn, does not appear in this equation, it means that $x_j$ does not directly participate in $\mathcal{M}$.

In the remainder of this paper, we will use matrix notation for presence and absence of variables in equations within a system of simultaneous structural equations and will call such a matrix a structure matrix.

**Definition 1 (structure matrix)** *The structure matrix $A$ of a system $\mathcal{S}$ of $n$ simultaneous structural equations $e_1, e_2, \ldots, e_n$ with $n$ variables $x_1, x_2, \ldots, x_n$ is a square $n \times n$ matrix in which element $(ij)$ (row $i$, column $j$) is non-zero if and only if variable $x_j$ participates in equation $e_i$. Non-zero elements of $A$ will be denoted by $X$ (a capital letter $X$) and zero elements by $0$.*

Note that the structure matrix is used for notational convenience only and does not mean that the discussion of simultaneous equations models is restricted to linear models.

**Example:** The following simple model, consisting of a set of simultaneous linear structural equations (along with the structure matrix), describes our perception of the interaction among the percentage of drunk drivers on the road ($d$), frequency of car accidents ($a$), and ratio of mortalities to the total number of people involved in these accidents ($m$).

$$\begin{cases} d = \mathcal{E}_1 \\ a + C_1 d = \mathcal{E}_2 \\ m + C_2 a = \mathcal{E}_3 \end{cases} \quad \begin{bmatrix} & m & a & d \\ (e_1) & 0 & 0 & X \\ (e_2) & 0 & X & X \\ (e_3) & X & X & 0 \end{bmatrix} \quad (3)$$

Note that each equation describes what we believe to be a mechanism. Drinking and accidents are involved in a mechanism — being under influence of alcohol interacts with driving abilities and effectively increases the likelihood of an accident (equation $e_2$). Mortality is involved in a mechanism with car accidents, but not with the percentage of drunk drivers (equation $e_3$). Our assumption here was that drinking is not involved in a direct functional relation with mortality. Further, as we believe that none of the variables in the model can affect $d$, we made it an exogenous variable (equation $e_1$). $C_1$ and $C_2$ are constants (positive or negative) and error variables are specified by a probability distribution. Note, that algebraically, this model is equivalent to the following (we preserved for the sake of demonstration the original constants and error terms):

$$\begin{cases} m + C_2 a + d = \mathcal{E}_1 + \mathcal{E}_3 \\ a + C_1 d = \mathcal{E}_2 \\ m + C_2 a = \mathcal{E}_3 \end{cases} \quad (4)$$

We do not consider this later model to be structural, because the first equation would suggest a single mechanism involving drinking, accidents, and mortality. This violates our view of the mechanisms operating in this system and is, therefore, not structural. Still, this model has the same numerical solution as the model in (3).    □

Simultaneous structural equations models have been a standard way of representing static systems in econometrics (Hood & Koopmans, 1953). Structural form is the most natural form of building a model — one composes a system modularly from pieces based on available knowledge about the system's components. Yet, the main advantage of having a structural model of a system is that it can aid predictions in the presence of changes in structure. We will end this section with a discussion of this important property of structural equations models.

It is easy to observe that simultaneous structural equations models imply asymmetric relations among variables. Consider the example model given in (3). A change in the error variable $\mathcal{E}_1$ will affect the value of $d$ directly, and the values of $a$ and $m$ indirectly. A change in $\mathcal{E}_3$, in turn, will affect only variable $m$ and leave all other variables unchanged.

A change in the structure of a system is modeled in a straightforward way in a simultaneous structural equations model: one alters or replaces the equations representing the modified mechanisms. Consider, for example, a policy that makes seat belts mandatory. We add a new variable $b$ standing for the belt usage (expressed, for example, by the ratio of the drivers who use belts after the policy has been imposed). Since the belt use is determined exogenously with respect to the system, we add an equation for $b$. By virtue of their design, it is reasonable to assume that seat belt usage interacts directly with accident mortality rate, hence, the mechanism that the new policy modifies is that described by the equation involving $m$.[5] The new

---

[5] If there were reasons to believe that seat belts usage

model will, therefore, take the following form:

$$\left\{\begin{array}{r}d = \mathcal{E}_1 \\ a + C_1 d = \mathcal{E}_2 \\ m + C_2 a + C_3 b = \mathcal{E}_4 \\ b = \mathcal{E}_5\end{array}\right. \quad \begin{array}{c} \\ (e_1) \\ (e_2) \\ (e_3) \\ (e_4) \end{array} \begin{bmatrix} m & a & d & b \\ 0 & 0 & X & 0 \\ 0 & X & X & 0 \\ X & X & 0 & X \\ 0 & 0 & 0 & X \end{bmatrix} \quad (5)$$

It follows from the modified version of the model that a change in $\mathcal{E}_5$ will only affect $b$ and then $m$. The values of $d$ and $a$ are uniquely determined by the first and second equations, hence, will remain unaffected by the change in structure. This agrees with our intuition that mandatory seat belts will not affect drivers' drinking habits and the number of accidents. If the model involved an alternative form of the equations, such as (4), we would have been in general unable to determine the effect of a change in the structure of the model. As it is impossible in such a form to identify the equations describing the altered mechanisms (note that in (4) $m$ and $a$ appear in two equations), it is not obvious which equations need to be modified and how.

## 2.2 CAUSAL ORDERING

This property of simultaneous structural equations models was made explicit by Simon (1953), who pointed out that interactions among variables in a self-contained simultaneous equations models are asymmetric and that this asymmetry leads to an ordering of the variables. He developed an algorithm for extracting this ordering and argued that, if each equation in the model is structural and each variable in the model that is assigned a constant value is an exogenous variable, then this ordering has a causal interpretation. Causal ordering is a mechanical procedure that retrieves the dependency structure in a set of simultaneous equations. This structure will correspond to the interactions in the real world in so far as the model corresponds to the real world.

The procedure of extracting the causal ordering from a simultaneous structural equations model works roughly as follows. A set of equations is *self-contained* if it has as many equations as variables and if every subset of equations has at least as many variables as equations. So a set of $n$ equations is self-contained if it contains $n$ unknowns and if every subset of $m$ equations has at least $m$ variables. Mechanisms in real-world systems often involve small number of elements, leading to structure matrices with many zeros. A set of structural equations will usually contain subsets that are self-contained (i.e., they also consist of as many equations as variables). A subset of $k$ equations with $k$ variables is called a subset of *degree k*. Simon proved that intersections of self-contained subsets are self-contained, thereby proving the existence of a minimal self-contained subset, i.e., one that does not have self-contained subsets (in the worst case, this

---

was involved in the mechanism that leads to an accident, we might have modified equation $e_2$ as well. Similarly, drinking might affect the probability of seat belt use and, hence, be implicated in the equation $e_4$.

subset will be equal to the entire set). The procedure recursively identifies minimal self-contained subsets, solves them for the variables that they contain, and substitutes the obtained values for each occurrence of each variable in the remaining equations. Note that these variables are exogenous to the subsets identified at a later stage. The procedure stops when no self-contained subsets can be identified. A self-contained subset is defined to be of *order k* if it was identified and solved at step $k$ of this procedure. The order in which the subsets were identified defines the causal ordering of the equations and variables. Variables exogenous to a subset are the direct causal predecessors of the variables in that subset. It is possible to construct a causal graph of the system by representing each variable in the system by a node and drawing directed arcs to each node from its direct causal predecessors. For the formal description of the procedure, see (Simon, 1953).

**Example:** The causal ordering procedure applied to the model described by (5) will first identify equations $e_1$ and $e_4$ as two self-contained structures of degree one. Both equations contain only one variable and, hence, are minimal subsets of zero order. There are no other such subsets. Solving $e_1$ for $d$ and $e_4$ for $b$, and substituting these values in $e_2$ and $e_3$, yields one self-contained structure of degree one, notably equation $e_2$. Since we are in step one of the procedure, $e_2$ is an equation of first order. Solving $e_2$ for $a$ and substituting this value in $e_3$, we are left with a single equation as the minimal self-contained subset of second order. The resulting causal ordering is:

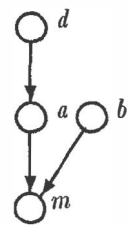

□

Causal ordering is an asymmetric relation between variables, determined by a collection of mechanisms embedded in a system. It is defined formally in the context of models of real-world systems, whose primitives are equations describing mechanisms acting in these systems. Mechanisms, as opposed to causal connections, are usually perceptible and, hence, form a sound operational basis for the approach. But none of these mechanisms determines the causal ordering in isolation: causal ordering is a property of a whole system rather than of an individual equation. We will subsequently call this view of causality *mechanism-based* to reflect its reliance on the notion of mechanisms in defining the causal structure of a system.

Causal ordering is qualitative in nature, in the sense that it does not require full algebraic specifications of the equations in the model. Actually, knowledge of which variables in a model participate in which equa-tions is sufficient. This, in turn, is equal to the knowledge of whether an element of the structure matrix is non-zero. Actual values of the coefficients (including their signs) and the algebraic form of the equations can remain unspecified.

No scientist will claim that a model he or she has proposed is the true model of the real-world system and, in that sense, the causal structure explicated by the procedure of causal ordering is to a certain extent subjective. It is as good as the current state of knowledge, as the physical, chemical, or social laws, and as good as the real-world measurements that it is based on and the approximations that the scientist was willing to make. This subjectivity seems to be an irreducible property of models, but luckily a property that is comparable to the subjectivity of science.

A possible criticism of causal ordering might be that it adds nothing new: whatever it produces is already embedded in the equations. Causality, in particular, must have been encoded in each of the equations by the author of the model. This critique is misplaced, however, because there is nothing in a typical equation that would suggest asymmetry. Causal ordering of variables becomes apparent only when the equation is placed in context. For example, the only information captured by structural equations for a bridge truss might be the points through which it is connected with the rest of the bridge. It is the context, the equations describing the remaining components of the bridge that will determine the role of the truss in the causal structure of the bridge and the direction of causality. The direction of the causal relation in one system can be easily reversed in another system. What causal ordering accomplishes is to explicate the asymmetry of the relations among variables in a simultaneous structural equations model once such a context has been provided.

The work on causal ordering originated in econometrics, where it was initially shown in the context of deterministic linear models (Simon, 1953). It was demonstrated to apply equally well to logical equations models (Simon, 1952) and linear models with error variables (Simon, 1954). It was shown to provide an interesting basis for treatment of the counterfactual conditional (Simon & Rescher, 1966). Recently, the method has been extended to non-linear and dynamic models, involving first-order differential equations (Iwasaki, 1988) and was shown to provide a sound basis for qualitative physics (Iwasaki & Simon, 1986) and non-monotonic reasoning (Simon, 1991).

## 3   CAUSALITY IN BAYESIAN BELIEF NETWORKS

It is often the case that, although something is known about the qualitative and statistical properties of a system's mechanisms, the exact functional form of the system's interactions is unknown. BBN models represent all interactions among a system's variables by means of probability distributions and, therefore, supply a way to model such cases.

The pure mathematical formalism of BBNs is based on factorization of the joint probability distribution of all variables in the model. Since this factorization is usually not unique, many equivalent models can be used to represent the same system, just as was the case with the simultaneous equations models. Models are strongly preferred that represent probabilistic independences explicitly in their graphical structure. Such models minimize the number of arcs in the graph, which in turn increases clarity and offers computational advantages.

Historically, BBN models were developed to represent a subjective view of a system elicited from a decision maker or a domain expert (Howard & Matheson, 1984). Although there are several empirically tested model-building heuristics, there are no formal foundations and the process is still essentially an art. Decision makers are usually encouraged to specify variables that are directly relevant probabilistically (or causally) to a variable and influence that variable directly. These variables neighbor one another in the graph and a directed arc is drawn between them. Often, the direction of this arc reflects the direction of causal influence, as perceived by the decision maker. Sometimes, the direction of the arc reflects simply the direction in which the elicitation of conditional probabilities was easier.

While it is certainly not the case that every directed arc in a BBN denotes causality, the formalism is capable of representing asymmetry among variables and, thereby, causality. This section examines the conditions under which one can reasonably interpret the structure of a BBN as a causal graph of the system that it represents. We will approach the problem of specifying these conditions by comparing BBNs to structural equations models. Our intention is not to replace BBNs with structural equations models, but to integrate the existing body of work on modeling natural systems, structure, and causality.

The argument contained in this section consists of three steps. First, we demonstrate that BBN models can be represented by simultaneous equations models, that is, that the joint probability distribution represented by any BBN model $B$ can also be represented by a simultaneous equations model $S$ (Theorem 1). We then show that the structure of $B$ is equivalent to the structure of a causal graph of $S$ obtained by the method of causal ordering (Theorem 2). But the structure of $B$ reflects the causal structure of the underlying system if and only if the structural model of that system shares the structure of $S$ (Theorem 3). So, we can reduce the semantic constraints on the structure of $B$ to the constraints on the structure of $S$.

The following theorem demonstrates that the joint probability distribution over $n$ variables of a BBN can be represented by a model involving $n$ simultaneous equations with these $n$ variables and $n$ additional independently distributed latent variables. We prove this theorem for discrete probability distributions, such as those represented by variables in BBNs. Intuitively, it appears that this theorem should extend to continuous distributions, although we leave the problem of demonstrating that this is indeed the case open.

**Theorem 1 (representability)** *Let $\mathcal{B}$ be a BBN model with discrete random variables. There exists a simultaneous equations model $\mathcal{S}$, involving all variables in $\mathcal{B}$, equivalent to $\mathcal{B}$ with respect to the joint probability distributions over its variables.*

**Proof:** The proof is by demonstrating a procedure for constructing $\mathcal{S}$. A BBN is a graphical representation of a joint probability distribution over its variables. This joint probability distribution is a product of the individual probability distributions of each of the variables. It is, therefore, sufficient to demonstrate a method for reproducing the probability distribution of each of the variables in $\mathcal{B}$. For the sake of simplicity, the proof is for BBNs with binary variables. Extension to discrete variables with any number of outcomes is straightforward. The outcomes of a variable $x$ will be denoted by $X$ and $\overline{X}$. For the sake of brevity, we will use $Pr(X)$ to denote $Pr(x = X)$.

We will construct one equation for each of the variables. Each equation will include an independent, continuous latent variable $\mathcal{E}$, uniformly distributed over the interval $[0,1]$. Note that $\forall x \ (0 < x \leq 1) \ Pr(\mathcal{E} \leq x) = x$. We start with an empty set $\mathcal{S}$ and then, for each variable $y$ in $\mathcal{B}$, we add one equation to $\mathcal{S}$ in the following way.

If $y$ has no predecessors, then the probability distribution of its outcomes is the prior distribution, $Pr(Y)$, $Pr(\overline{Y})$. The following deterministic equation with a latent variable $\mathcal{E}$ reproduces the distribution of $y$:

$$f_y(\mathcal{E}) = \begin{cases} Y & \text{if } \mathcal{E} \leq Pr(Y) \\ \overline{Y} & \text{if } \mathcal{E} \leq Pr(\overline{Y}) \end{cases}$$

If $y$ does have direct predecessors $x_1, x_2, \ldots, x_n$, each of the variables $x_i$ ($1 \leq i \leq n$) having outcomes $X_i$ and $\overline{X_i}$, then its probability distribution is a distribution conditional on all possible outcomes of these predecessors (values $\mathcal{E}_i$ are introduced for the sake of brevity in future references to individual conditional probabilities).

$$Pr(Y|X_1, X_2, \ldots, X_n) = \mathcal{E}_1$$
$$Pr(Y|\overline{X_1}, X_2, \ldots, X_n) = \mathcal{E}_2$$
$$Pr(Y|X_1, \overline{X_2}, \ldots, X_n) = \mathcal{E}_3$$
$$Pr(Y|\overline{X_1}, \overline{X_2}, \ldots, X_n) = \mathcal{E}_4$$
$$\ldots$$
$$Pr(Y|\overline{X_1}, \overline{X_2}, \ldots, \overline{X_n}) = \mathcal{E}_{2^n}$$

The following deterministic equation with the latent variable $\mathcal{E}$ reproduces the distribution of $y$:

$$f_y(x_1, x_2, \ldots, x_n, \mathcal{E}) =$$

$$= \begin{cases} Y & \begin{array}{l}\text{if } x_1 = X_1, x_2 = X_2, \ldots, x_n = X_n, \mathcal{E} \leq \mathcal{E}_1 \\ \text{or } x_1 = \overline{X_1}, x_2 = X_2, \ldots, x_n = X_n, \mathcal{E} \leq \mathcal{E}_2 \\ \text{or } x_1 = X_1, x_2 = \overline{X_2}, \ldots, x_n = X_n, \mathcal{E} \leq \mathcal{E}_3 \\ \text{or } x_1 = \overline{X_1}, x_2 = \overline{X_2}, \ldots, x_n = X_n, \mathcal{E} \leq \mathcal{E}_4 \\ \ldots \\ \text{or } x_1 = \overline{X_1}, x_2 = \overline{X_2}, \ldots, x_n = \overline{X_n}, \mathcal{E} \leq \mathcal{E}_{2^n} \end{array} \\ \\ \overline{Y} & \begin{array}{l}\text{if } x_1 = X_1, x_2 = X_2, \ldots, x_n = X_n, \mathcal{E} > \mathcal{E}_1 \\ \text{or } x_1 = \overline{X_1}, x_2 = X_2, \ldots, x_n = X_n, \mathcal{E} > \mathcal{E}_2 \\ \text{or } x_1 = X_1, x_2 = \overline{X_2}, \ldots, x_n = X_n, \mathcal{E} > \mathcal{E}_3 \\ \text{or } x_1 = \overline{X_1}, x_2 = \overline{X_2}, \ldots, x_n = X_n, \mathcal{E} > \mathcal{E}_4 \\ \ldots \\ \text{or } x_1 = \overline{X_1}, x_2 = \overline{X_2}, \ldots, x_n = \overline{X_n}, \mathcal{E} > \mathcal{E}_{2^n} \end{array} \end{cases}$$

The above demonstrates that the value of any node in a BBN can be expressed by a deterministic function of the values of all its direct predecessors and a single independently distributed latent variable. For a BBN with $n$ nodes, we have constructed a self-contained set of $n$ simultaneous equations with $n$ variables and $n$ independent uniformly distributed continuous latent variables. The probability distribution of each variable in $\mathcal{S}$ is identical to the distribution of a corresponding node in $\mathcal{B}$. This makes $\mathcal{S}$ equivalent to $\mathcal{B}$ with respect to the joint probability distribution of the variables. □

The construction of an equivalent simultaneous equations model $\mathcal{S}$ for a BBN $\mathcal{B}$, outlined in the above proof, is rather straightforward. The goal is to describe each element of the conditional probability matrix of a node $y$ in $\mathcal{B}$. Each logical condition on the right-hand side of the equations specifies one element of this matrix by listing a combination of outcomes of parents of $y$. The exact numerical value of the conditional probability for that element is then given by the probability of an event involving the latent variable $\mathcal{E}$.

**Example:** Let $\mathcal{B}$ be a BBN with nodes $x$ and $y$.

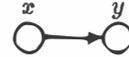

Let the distribution of $x$ be $Pr(X) = 0.4$, $Pr(\overline{X}) = 0.6$, and the conditional distribution of $y$ be $Pr(Y|X) = 0.7$, $Pr(Y|\overline{X}) = 0.2$, $Pr(\overline{Y}|X) = 0.3$, $Pr(\overline{Y}|\overline{X}) = 0.8$.

A simultaneous equations model for $\mathcal{B}$ is:

$$\begin{cases} f(\mathcal{E}_x) = \begin{cases} X & \text{if } \mathcal{E}_x \leq 0.4 \\ \overline{X} & \text{if } \mathcal{E}_x \leq 0.6 \end{cases} \\ f(x, \mathcal{E}_y) = \begin{cases} Y & \text{if } x = X, \mathcal{E}_y \leq 0.7 \text{ or } x = \overline{X}, \mathcal{E}_y \leq 0.2 \\ \overline{Y} & \text{if } x = X, \mathcal{E}_y \leq 0.3 \text{ or } x = \overline{X}, \mathcal{E}_y \leq 0.8 \end{cases} \end{cases}$$

□

BBNs are acyclic, which excludes self-contained structures of degree higher than one. It is, therefore, obvious that the converse of Theorem 1 is not true. For example, models with feedback loops cannot be represented by BBNs.

The following theorem establishes an important property of a structural equations model of a system with an assumption of causal acyclicity. This property implies that the structure obtained by the method of causal ordering from a structural equations model $S$ constructed in the proof of Theorem 1 reflects the structure of the equivalent BBN $\mathcal{B}$.

**Theorem 2 (acyclicity)** *The acyclicity assumption in a causal graph corresponding to a self-contained system of equations $S$ is equivalent to the following condition on $S$: Each equation $e_i \in S : f(x_1, \ldots, x_n, \mathcal{E}_i) = 0$ forms a self-contained system of some order $k$ and degree one, and determines the value of some argument $x_j$ $(1 \leq j \leq n)$ of $f$, while the remaining arguments of $f$ are direct predecessors of $x_j$ in causal ordering over $S$.*

**Proof:** $\boxed{\Longrightarrow}$ Acyclicity, according to the procedure of causal ordering, means that in the process of extracting the causal ordering from $S$, there is no self-contained structure of degree higher than one (i.e., containing more than one equation). We will show that given the assumption of acyclicity, the structure matrix $A$ of the equations in $S$ is triangular. Then, by the considerations analogous to Gaussian elimination and by causal ordering the theorem follows.

We will transform $A$ into a lower-triangular matrix by a series of operations involving row interchanges and column interchanges. Both operations preserve the causal ordering of variables in $S$: row interchange is equivalent to changing the order of equations in $S$; column interchange is equivalent to renaming the variables in $S$. Both the order of equations and names of variables are insignificant and do not depend on the functional form of equations. We will work along the diagonal from the upper-left to the lower-right corner and always rearrange rows below and columns to the right of the current diagonal element.

Since all self-contained structures in $S$ are all of degree one, we know in the beginning (row 0 and column 0) that there will be at least one equation containing only one variable. Suppose row $i$ describes the coefficients of such an equation. We know that there will be only one non-zero coefficient in this equation. If this coefficient is located in column $j$, we interchange row 0 with row $i$ and column 0 with column $j$. Now, the first element on the diagonal (first pivot) will contain a non-zero; all other elements in row 0 will be zeros. Now we proceed with the next pivot. Processing the k-th pivot, we are by assumption guaranteed that in the sub-matrix $[k:n; k:n]$ there will be at least one self-contained structure of degree one, which means that there will be at least one row with only one non-zero element. Suppose, it is row $i$ and the non-zero element is in column $j$. We then interchange row $k$ with row $i$ and column $k$ with column $j$. Since the $k$-th pivot is the only non-zero element in the current $k$-order self-contained structure, all elements to the right of it are zeros. Note also that this interchange does not affect the zero elements above diagonal in the rows 0 to $k-1$, since all columns from $k$ to $n$ had their elements 0 to $k-1$ equal to zero.

By the considerations based on Gaussian elimination, each of the diagonal elements $a_{ii}$ is the coefficient of some variable $x_i$, determined by the equation $e_i$. Each of the other non-zero elements left of $a_{ii}$ denotes presence in the equation $e_i$ of a variable that is determined before $x_i$, that is a direct predecessor of $x_i$ in the causal ordering over $S$.

$\boxed{\Longleftarrow}$ If each equation determines exactly one variable, it follows that at each level in the procedure of extracting the causal ordering, there are only self-contained structures of degree one, which in turn guarantees acyclicity of the causal graph. □

Each equation in the simultaneous equations model $S$ constructed in the proof of the Theorem 1 involves a node in the graph representing the BBN $\mathcal{B}$ and all its immediate predecessors. By Theorem 2, the causal graph of $S$ derived by the method of causal ordering will have the same structure as $\mathcal{B}$. This observation and its implications are formalized in the following theorem.

**Theorem 3 (causality in BBNs)** *A Bayesian belief network $\mathcal{B}$ reflects the causal structure of a system if and only if (1) each node of $\mathcal{B}$ and all its direct predecessors describe variables involved in a separate mechanism in the system\* and (2) each node with no predecessors represents an exogenous variable.\*\**

**Proof:** By Theorem 1, there exists a simultaneous equations system $S$ that is equivalent to $\mathcal{B}$. Each equation in $S$ involves a node of $\mathcal{B}$ and all its direct predecessors. We know that $\mathcal{B}$ is acyclic, so by Theorem 2 the structure of $\mathcal{B}$ is equivalent to the structure of a causal graph obtained by the method of causal ordering from $S$.

By the assumptions underlying causal ordering, $\mathcal{B}$ reflects the causal structure of the underlying system if and only if $S$ is a structural model of that system, i.e., if each of its equations is a structural equation and each of its exogenous variables is a true exogenous variable. This is what the theorem states. □

## 4 CONCLUSION

Knowledge of causal asymmetries in a system is necessary in predicting the effects of changes in the structure of the system and, because of the role of causality in human reasoning, is essential in human-computer interfaces to decision support systems. Although many researchers refer to the concept of causality, there seems to be no consensus as to what causality in BBN models means and how BBNs' directed arcs should be interpreted. We reviewed the mechanism-based view of causality in structural equations models and we have shown that it is applicable to BBN models. We have explicated the conditions that need to be satisfied in order for a BBN to be a causal model.

Theorem 3 demonstrates that directed arcs in BBNs play a role that is similar in its representational power to the structure (in terms of the presence or absence of variables in equations) of simultaneous equations models. We can view the graphical structure of a BBN as a qualitative specification of the mechanisms acting in a system. Similarly to the mathematical transformations on structural equations models (such as row combination in linear models), we can obtain BBNs that are equivalent with respect to the probability distribution of its variables by reversing network's arcs. However, similarly to simultaneous equations models, such transformations will lead to loss of structural information. There is only one graphical structure that fulfills the semantic requirements stated in the theorem and can be given a causal interpretation.

---

\* i.e., $\forall i \; x_i = f_i(\pi_i, \mathcal{E}_i)$

\*\* vacuous: any variable can be so, conceptually

Our analysis shows how prior theoretical knowledge about a domain, captured in structural equations, can aid construction of BBNs. Given the assumption of acyclicity, an equation involves a node and all its direct predecessors, as shown in Theorem 2.. This provides valuable information about adjacencies in the constructed network. Currently, both, the structure and the numerical probability distributions in BBNs are elicited from a human expert and are a reflection of the expert's subjective view of a real world system. Existing theoretical knowledge, if incorporated at the model building stage, should aid human experts, make model building easier, and, finally, improve the quality of constructed models.

### Acknowledgments

We thank anonymous reviewers for insightful remarks.